%% file: main.tex
\documentclass[letter, 10pt, conference]{ieeeconf}  
\usepackage[cmex10]{amsmath}
\usepackage{mathtools}
\usepackage{tabularx,booktabs}
\usepackage{graphicx}
\usepackage{array}
\usepackage[utf8]{inputenc}
\usepackage{tabu}
\usepackage{tabularx}
\usepackage{booktabs}
\usepackage[english]{babel}
\usepackage{csquotes}
\usepackage[skip=2pt,font=small]{caption}
\usepackage{footnote}
\usepackage{dblfloatfix}
\usepackage{amssymb}
\usepackage{verbatim}
\usepackage{multirow}
\usepackage{dsfont}

\newcommand\newblock{\hskip .11em\@plus.33em\@minus.07em}

\def\@fnsymbol#1{\ensuremath{\ifcase#1\or *\or \dagger\or \ddagger\or
   \mathsection\or \mathparagraph\or \|\or **\or \dagger\dagger
   \or \ddagger\ddagger \else\@ctrerr\fi}}
\usepackage{url}
\usepackage[square,numbers,comma, sort]{natbib}
\RequirePackage{color}
\usepackage[colorlinks]{hyperref}
\ifdefined\nohyperref\else\ifdefined\hypersetup
  \definecolor{mydarkblue}{rgb}{0,0.08,0.45}
  \hypersetup{ %
    pdftitle={},
    pdfauthor={},
    pdfsubject={},
    pdfkeywords={},
    pdfborder=0 0 0,
    pdfpagemode=UseNone,
    colorlinks=true,
    linkcolor=mydarkblue,
    citecolor=mydarkblue,
    filecolor=mydarkblue,
    urlcolor=magenta,
    pdfview=FitH}
\DeclareQuoteStyle[american]{english}
    {\itshape\textquotedblleft}
    [\textquotedblleft]
    {\textquotedblright}
    [0.05em]
    {\textquoteleft}
    {\textquoteright}
\DeclareUnicodeCharacter{2217}{-}
\graphicspath{ {images/} }
\usepackage[linesnumbered, ruled]{algorithm2e}

\IEEEoverridecommandlockouts                              

\renewcommand{\footnoterule}{%
  \hrule width 2in
  \kern 2pt
}

\definecolor{mycolor}{RGB}{147,112,219}

\begin{document}

\title{\LARGE \bf Hierarchical Cross-Modal Agent for \\Robotics Vision-and-Language Navigation}
\author{Muhammad Zubair Irshad\textsuperscript{*}, Chih-Yao Ma\textsuperscript{*}\textsuperscript{$\dagger$}, Zsolt Kira\textsuperscript{*}
}

\twocolumn[{%
\renewcommand\twocolumn[1][]{#1}%
\maketitle
\begin{center}
    \centering
    \includegraphics[width=0.98\textwidth]{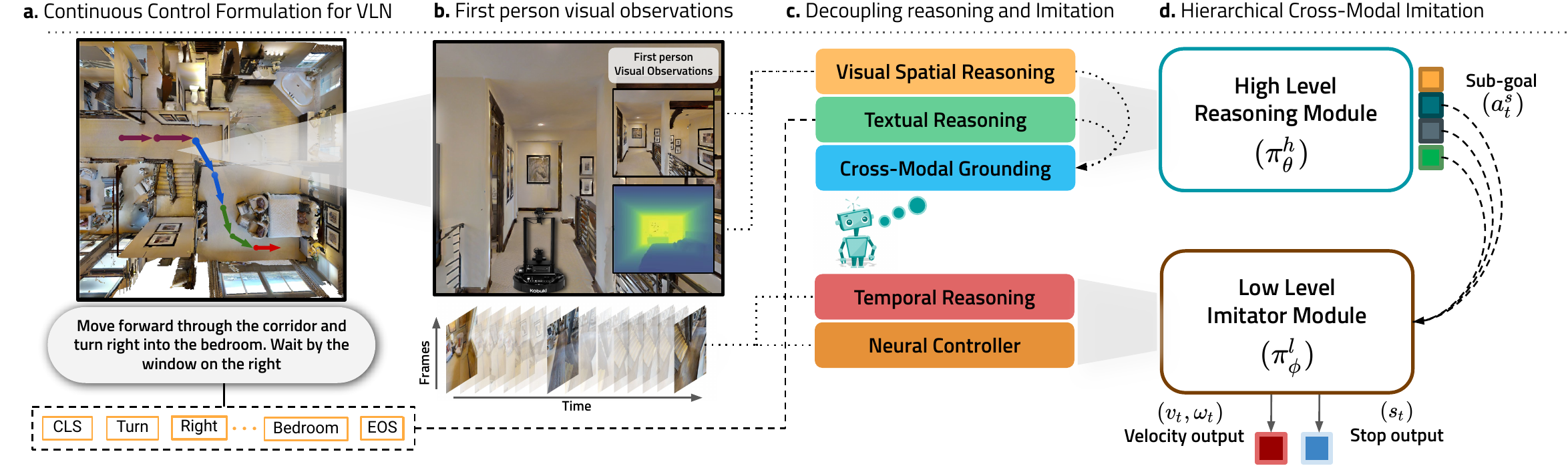}
    \captionsetup{width=\linewidth}
    \captionof{figure}{
    \textbf{Overview:} Robotics Vision-and-Language Navigation (Robo-VLN) task in continuous environments and our proposed Hierarchical Cross-Modal (HCM) agent. The agent decouples reasoning and imitation through a modularized training regime to solve the complex long-horizon Robo-VLN task.
    }
   \label{overview}
\end{center}%
}]
{
  \renewcommand{\thefootnote}%
    {\fnsymbol{footnote}}
  \footnotetext[1]{Georgia Institute of Technology \tt\scriptsize (mirshad7,zkira)@gatech.edu}
    \footnotetext[2]{Now at Facebook \tt\scriptsize cyma@fb.com}
}

\begin{abstract}
Deep Learning has revolutionized our ability to solve complex problems such as Vision-and-Language Navigation (VLN). This task requires the agent to navigate to a goal purely based on visual sensory inputs given natural language instructions. However, prior works formulate the problem as a navigation graph with a discrete action space. In this work, we lift the agent off the navigation graph and propose a more complex VLN setting in continuous 3D reconstructed environments. Our proposed setting, Robo-VLN, more closely mimics the challenges of real world navigation. Robo-VLN tasks have longer trajectory lengths, continuous action spaces, and challenges such as obstacles. We provide a suite of baselines inspired by state-of-the-art works in discrete VLN and show that they are less effective at this task. We further propose that \textit{decomposing the task} into specialized high- and low-level policies can more effectively tackle this task. With extensive experiments, we show that by using layered decision making, modularized training, and decoupling reasoning and imitation, our proposed Hierarchical Cross-Modal (HCM) agent outperforms existing baselines in all key metrics and sets a new benchmark for Robo-VLN.
\end{abstract}


%

\section{Introduction}

The promise of personal assistant robots that can seamlessly follow human instructions in real life environments has long been sought after. Recent advancements in deep learning (to extract meaningful information from raw sensor data) and deep reinforcement learning (to learn effective decision-making policies) have enabled some progress towards this goal~\cite{vasudevan2020talk2nav, majumdar2020improving, hao2020towards}. Due to the difficulty of collecting data in these contexts, a great deal of work has been done using photo-realistic simulations such as those captured through Matterport3D panoramas in homes~\cite{Matterport3D} or point-cloud meshes in Gibson~\cite{gibsonenv}. For example, a number of works have investigated autonomous agents that can follow rich, natural-language instructions in such simulations~\cite{ma2019theregretful, Wang2019ReinforcedCM, DBLP:conf/nips/FriedHCRAMBSKD18}. Precisely defined, Vision-and-Language Navigation (VLN) is a task which requires the agent to navigate to a goal location purely based on visual inputs and provided instructions in the absence of a prior global map~\cite{mattersim}.

While increasingly effective neural network architectures have been developed for these tasks, many limitations still exist that prevent their applicability to real-world robotics problems. Specifically, previous works~\cite{Wang2019ReinforcedCM, DBLP:conf/nips/FriedHCRAMBSKD18,46942, Matterport3D, Zhuetal} have focused on a simpler subset of this problem by defining the instruction-guided robot trajectories as either a discrete navigation graph~\cite{Matterport3D, mattersim} or assuming the action space of the autonomous agent comprises of discrete values~\cite{ALFRED20, krantz2020navgraph}. These formulations assume known topology, perfect localization and deterministic navigation from one viewpoint to the next in the absence of any obstacles~\cite{krantz2020navgraph}. Hence these assumptions significantly deviate from the real world both in terms of control and perception.
 
\begin{table*}[t!]
\scriptsize
\centering
\vspace{+0.15cm}
\caption{\textbf{Comparison between our proposed Robo-VLN setting and prior environments used for Vision-and-Language Navigation}} \label{tab:a}
\label{rvln_comparison}
\resizebox{0.9\textwidth}{!}{%
\begin{tabular}{c*{8}{>{$}c<{$}}}
\toprule

& \multicolumn{3}{c}{\textbf{---Simulation---}} &\multicolumn{2}{c}{\textbf{---Environment---}}&\multicolumn{2}{c}{\textbf{---Instructions---}} \\
\cline{1-4}\cline{4-6}\cline{6-8} \\ 
\textbf{} & \textbf{Action space}& \textbf{Granularity} & \textbf{Agent} & \textbf{Navigation} & \textbf{Type}  & \textbf{Richness} & \textbf{Generation} \\
\midrule
\textbf{Touchdown~\cite{ai2thor}, R2R~\cite{mattersim}} & $Discrete$ & $High$ & $Virtual$ & $Unconstrained$ & $Photo-realistic$ & $Complex$ & $Human-annotated$\\

\textbf{Follow-net~\cite{46942}}  & $Discrete$ & $High$ & $Virtual$ & $Constrained$ & $Synthetic$ & $Simple$ & $Human-annotated$\\

\textbf{LANI~\cite{misra-etal-2018-mapping}}  & $Discrete$ & $High$ & $Virtual$ & $Constrained$ & $Synthetic$ & $Simple$ & $Template based$\\

\textbf{VLN-CE~\cite{krantz2020navgraph}} & $Discrete$ & $High$ & $Virtual$ & $Unconstrained$ & $Photo-realistic$ & $Complex$ & $Human-annotated$\\

\midrule
\textbf{Robo-VLN (Ours)}  & $Continuous$ & $High/Low$ & $Robotics$ & $Unconstrained$ & $Photo-realistic$ & $Complex$ & $Human-annotated$ \\
\bottomrule
\end{tabular}
}
\end{table*}
As a first contribution, we focus on a richer VLN formulation which is defined in continuous environments over long horizon trajectories. Our proposed setting, \textbf{Robo-VLN} (\textbf{Robo}tics \textbf{V}ision-and-\textbf{L}anguage \textbf{N}avigation), is summarized in Figure~\ref{overview} and Section~\ref{robo-vln}. We lift the agent off the navigation graph, making the language guided navigation problem richer, more challenging, and closer to the real world. 

In an attempt to solve the language-guided navigation (VLN) problem, recent learning-based approaches~\cite{ma2019theregretful, ma2019selfmonitoring, wang2019reinforced} make use of sequence-to-sequence architectures~\cite{10.5555/2969033.2969173}. However, when tested for generalization performance in unseen environments, these approaches (initially developed for shorter horizon nav-graph problems) translate poorly to more complex settings~\cite{ALFRED20,krantz2020navgraph}, as we also showed for Robo-VLN in Section~\ref{experiments}.
Hence, for our proposed continuous VLN setting over long-horizon trajectories, we present an approach utilizing \textit{hierarchical decomposition}. Our proposed method leverages hierarchy to decouple cross-modal reasoning and imitation, thus equipping the agent with the following key abilities: 

\textbf{1. Decouple Reasoning and Imitation.}
The agent is comprised of a high-level policy and a corresponding low-level policy. The high-level policy is tasked with aligning the relevant instructions with observed visual cues as well as reasoning over which instructions have been completed, hence producing a sub-goal output through cross-modal grounding. 
The low-level policy imitates the feedback controller based on sub-goal information and observed visual states. A layered decision making allows spatially different reasoning at different levels in the hierarchy, hence specializing each policy with a dedicated reasoning abstraction level.

\textbf{2. Modularized Training.}
Disentangling reasoning and controls allows fragmenting a complex long horizon problem into shorter time horizon problems.
Since each policy is tasked with fulfilling a dedicated goal, each module utilizes separate end-to-end training with sparse communication between the hierarchy in terms of sub-goal information.
In summary, we make the following contributions: 
\begin{itemize}

\item To the best of our knowledge, we present the first work on formulating Vision-and-Language Navigation (VLN) as a continuous control problem in photo-realistic simulations, hence lifting the agent of the assumptions enforced by navigation graphs and discrete action spaces.

\item We formulate a novel hierarchical framework for Robo-VLN, referred to as \textbf{H}ierarchical \textbf{C}ross-\textbf{M}odal Agent (\textbf{HCM}) for effective attention between different input modalities through a modularized training regime, hence tackling a long-horizon and cross-modal task using layered decision making.
\item Provide a suite of baseline models in Robo-VLN inspired by recent state-of-the-art works in VLN and present a comprehensive comparison against our proposed hierarchical approach --- Our work sets a new strong benchmark performance for a long horizon complex task, Robo-VLN, with over 13\% improvement in absolute success rate in unseen validation environments.

\end{itemize} 

\section{Related Work} \label{sec:A}
\textbf{Vision-and-Language Navigation.} 
Learning based navigation has been explored in both synthetic~\cite{DBLP:journals/corr/KempkaWRTJ16, DBLP:journals/corr/abs-1712-05474, wu2018building} and photo-realistic~\cite{habitat19iccv, Matterport3D, gibsonenv} environments. For a navigation graph based formulation of the VLN problem (i.e. discrete action space), previous works have utilized hybrid reinforcement learning~\cite{DBLP:conf/eccv/WangXWW18}, behavior cloning~\cite{eqa_modular}, speaker-follower~\cite{NIPS2018_7592} and sequence to sequence based approaches~\cite{mattersim}. Subsequent methods have focused on utilizing auxiliary losses~\cite{ma2019selfmonitoring, Zhu_2020_CVPR}, backtracking~\cite{ma2019theregretful} and cross-modal attention techniques~\cite{DBLP:conf/cvpr/WangHcGSWWZ19,DBLP:journals/corr/abs-1905-13358,landi2020perceive} to improve the performance of VLN agents. Our work, in contrast to discrete VLN setting \cite{krantz2020navgraph,mattersim} (see Table \ref{rvln_comparison}), focuses on a much richer VLN formulation, which is defined for continuous action spaces over long-horizon trajectories. 
We study the new continuous Robo-VLN setting and propose hierarchical cross-modal attention and modularized training regime for such task.

\textbf{Hierarchical Decomposition.}
Hierarchical structure is most commonly utilized in the context of Reinforcement Learning over long-time horizons to improve sample efficiency~\cite{Sutton:1999, Sutton99betweenmdps, Vezhnevets2017FeUdalNF}. Our work closely relates to the options framework in Reinforcement Learning~\cite{,eqa_modular, pmlr-v80-le18a, Sutton:1999, pmlr-v54-fruit17a} where the top-level policy identifies high-level decisions to be fulfilled by a bottom-level policy. In relation to other works which utilize sub-task decomposition for behaviour cloning~\cite{pmlr-v80-le18a, Roh2019Conditional}, we show that decomposing hierarchy based on reasoning and imitation are quite effective for long-horizon multi-modal tasks such as Robo-VLN. 

\section{Robotics Vision-and-Language Navigation Environment (Robo-VLN)}\label{robo-vln}

Different from existing VLN environments, we propose a new continuous environment for VLN that more closely mirrors the challenges of the real world, Robo-VLN --- a continuous control formulation for Vision-and-Language Navigation. 
Compared to navigation graph based~\cite{mattersim} and discrete VLN settings~\cite{krantz2020navgraph}, Robo-VLN provides longer horizon trajectories (4.5x average number of steps), more visual frames ($\sim$3.5M visual frames), and a balanced high-level action distribution (see Figure~\ref{comparison_discrete}). Hence, making the problem more challenging and closer to the real-world.

\subsection{Problem Definition}
Formally, consider an autonomous agent $\tilde{\mathcal A}$ in an unknown environment $\tilde{\mathcal E}$. 
The goal of a Robo-VLN agent is to learn a policy $a_{t} = \pi(x_{t}, q_{t}, \theta)$ where the agent receives visual observations ($x_{t}$) from the environment $\tilde{\mathcal E}$ at each time-step ($t$) while following a provided instruction ($q$) to navigate to a goal location $\mathcal G$. 
$\theta$ denotes the learnable parameters of the policy $\pi$. 
The action space of the agent consists of continuous linear and angular velocity ($v_{t}, \omega_{t}$) and a discrete stop action ($s_{t}$). 
An episode ($\tau$) is considered successful if agent's distance to the goal is less than a threshold ($d_{a}<3m$) and the agent comes to a stop by either taking the stop action ($s_{t}$) or decreasing its angular velocity below a certain threshold.

\subsection{Constructing Continuous VLN in 3D Reconstructions}
To make the continuous VLN formulation possible in 3D reconstructed environments, we port over human annotated instructions ($q_{t}$) corresponding to sparse way-points ($z_{t}$) along each instruction-trajectory pair in Room2Room (R2R) dataset \cite{mattersim}, using a continuous control formulation. We do this in 2 stages as follows: 

\begin{figure}[t!]
\centering
\includegraphics[width=\columnwidth]{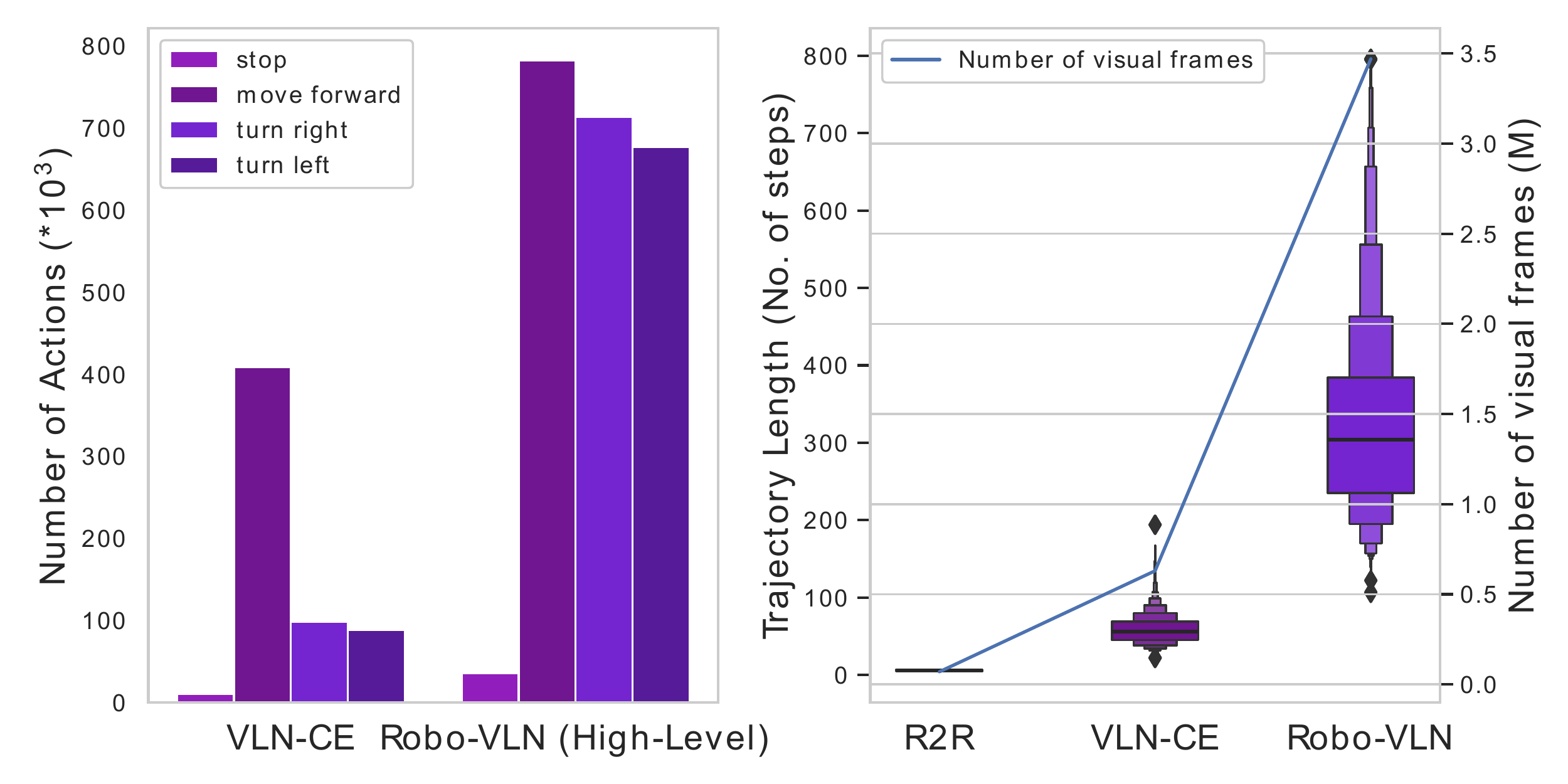}
\centering
  \caption{
  \textbf{Robo-VLN compared with discrete VLN settings:} VLN-CE~\cite{krantz2020navgraph} and R2R~\cite{mattersim}.
  We provide longer horizon trajectories (4.5x average number of steps, over 3M visual frames, and a balanced high-level action distribution.
  }
  \label{comparison_discrete}
\end{figure}

\textbf{Ground-truth oracle feedback controller in 3D reconstructed environments.}
We consider the robotic agent to be a differential drive mobile robot, Locobot~\cite{pyrobot2019}, with a specified radius and height. We develop $A^{\star}$ planner to compute high-level oracle actions ($a^{h}_{t}$) along the shortest path to the goal and use a feedback controller~\cite{9780133496598} to convert the discrete R2R trajectories \cite{mattersim} into continuous ones.
The low-level oracle controller ($u_{t}$) outputs velocity commands $(v_{t}, \omega_{t})$ given sparse way-points ($z_{t}$) along a given language-guided navigation trajectory from the R2R dataset~\cite{Matterport3D}.
The converted continuous actions from the low-level controller will then be used as ground-truth low-level supervisions $a^{l}_{t}$ when training the navigation agents.
We create this continuous control formulation inside Matterport 3D environments~\cite{Matterport3D} by considering the Locobot robot as a 3D mesh inside 3D reconstructed environments (see Figure \ref{qualitative}).
We use the robot's dynamics~\cite{10.5555/1855026} to predict next state ($\hat{x}_{t+1}$) given current state ($\hat{x}_{t}$) and controller actions ($a^{l}_{t}$). 
Similar to Habitat~\cite{habitat19iccv}, we render the mesh for any arbitrary viewpoint by taking the position generated by the dynamic model inside the 3D reconstruction.

\begin{figure*}[t!]
\centering
\includegraphics[width=0.8\textwidth]{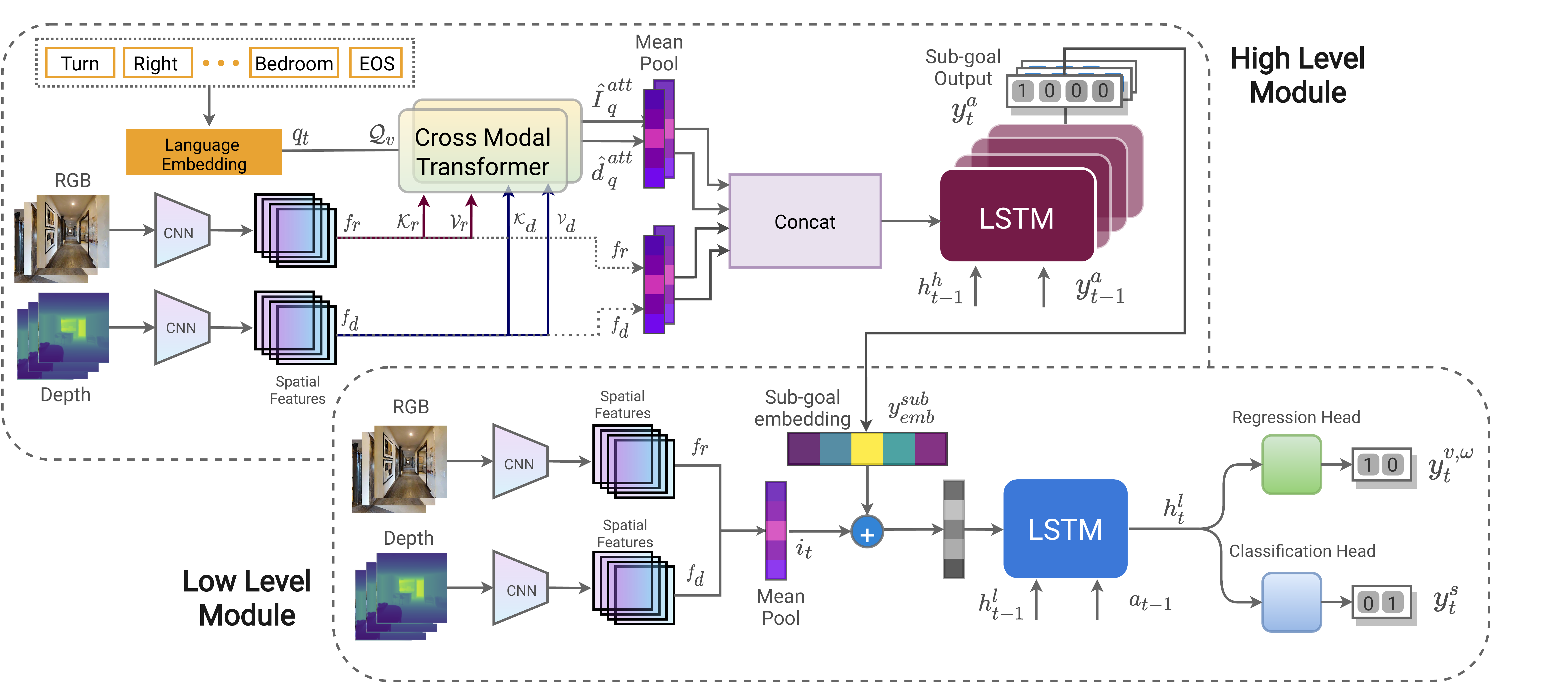}
\captionof{figure}{
\textbf{Hierarchical Cross-Modal Agent (HCM):} Our proposed agent consists of a \textit{high-level module} and a corresponding \textit{low-level module}. High-level module predicts the sub-goal output based on alignment between instructions and visual observations. Low-level module translates the high-level sub-goal output to linear and angular velocities using an imitation learning policy.}
\label{framework}
\end{figure*}

\textbf{Obtaining Navigable Instruction-Trajectory pairs.}
Given a feedback controller of the form $a^{l}_{t} = u_{t}(z_{t})$ and high-level sparse viewpoints ($z_{t} = [z_{1}, \ldots, z_{N}]$ along the language guided navigation trajectory inside a reconstructed mesh, we search for the navigable space $h_{nav}(z_{t})$ using collision detection.
We find navigable space for all the trajectories present in the R2R dataset~\cite{mattersim}.
This procedure ensures the transfer of only the navigable trajectories from R2R dataset to the continuous control formulation in Robo-VLN; hence, we eliminate non-navigable unrealistic paths for a mobile robot, such as climbing up the stairs and moving through obstacles. 
Through this approach, we transferred 71\% of the trajectories from the discrete VLN setting (VLN-CE~\cite{krantz2020navgraph}) while preserving all the environments in the Matterport3D dataset~\cite{Matterport3D}. At the end, Robo-VLN's expert demonstration provide first person RGB-D visual observations ($i_{t}$), human instructions ($q_{t}$), and oracle actions ($a^{h}_{t}, a^{l}_{t}$) for each instruction-trajectory pair.

\section{Hierarchical Cross-Modal Agent}
Learning an effective policy ($\pi$) for a long horizon continuous control problem entails preserving the temporal states as well as spatially reasoning about the surroundings. 
We therefore propose a hierarchical agent to tackle the Robo-VLN task as it effectively disentangles different dedicated tasks through layered decision making. 
Given states ($\mathcal{X}=\{x\}$) and instructions ($\mathcal{Q}=\{q\}$), our agent leverages these inputs and learns a high-level policy ($\pi^{h}_{\theta}:\mathcal{X} \times \mathcal{Q} \rightarrow \mathcal{A}_{s,t}$) and a corresponding low-level policy ($\pi^{l}_{\gamma}:\mathcal{X} \times \mathcal{A}_{s,t} \rightarrow \mathcal{A}_{l,t}$). 
The high-level policy consistently reasons about the alignment between input textual and visual modalities to produce a sub-goal output ($\mathcal{A}_{s,t}$). The low-level policy ensures that the high-level sub-goal is translated to low-level actions ($\mathcal{A}_{l,t}$) effectively by imitating the expert controller through an imitation learning policy. Our approach is summarized in Figure \ref{framework} and subsequent sections.

\subsection{High-Level Policy}
The high-level policy ($\pi^{h}_{\theta}$) decides a short-term goal ($a^{h}_{t}$) based on the input instructions ($q_{t}$) and observed visual information $x_{t} = \{ r_{t}, d_{t}\}$ from the environment at each time-step, where $r_{t}$, $d_{t}$ denote the RGB and Depth sensor readings respectively. $\pi^{h}_{\theta}$ consists of an encoder-decoder architecture with cross attention between the modules. Subsequent modules of the high-level policy ($\pi^{h}_{\theta}$) are described below.

\textbf{Multi-Modal Cross Attention Encoder.}
Given a natural language instruction comprised of $k$ words, we denote its feature representation as $\{q = q^{1}_{t}, q^{2}_{t}, \ldots, q^{k}_{t} \}$, where $q^{i}_{t}$ is the encoded feature representation of the $i_{th}$ word using BERT embedding~\cite{devlin-etal-2019-bert} to extract meaningful representation of words in the sentence.
To encode the observed RGB-D states ($r_{t}$ $\in$ $\mathds{R}^{h_{o}\times w_{o}\times 3 }$, $d_{t}$ $\in$ $\mathds{R}^{h_{o}\times w_{o}}$), we generate a low-resolution spatial feature representations $f_{r}$ $\in$ $\mathds{R}^{H_{s}\times W_{s}\times C_{s}}$ and $f_{d}$ $\in$ $\mathds{R}^{H_{s}\times W_{s}\times C_{s}}$ by using a pre-trained ConvNet backbone, where $H_{s}=W_{s} = 7$ and $C_{s}=2048$.
At each time-step $t$, we combine the individual RGB ($f_{r}$) and Depth ($f_{d}$) spatial features with encoded language representation ($q_{t}$) using a Transformer module~\cite{NIPS2017_7181}. Each Transformer module is comprised of stacked multi-head attention block $(\mathcal{A}_{M})$ followed by a position-wise feed-forward block. We utilize layer normalizations~\cite{ba2016layer} between these blocks along with the residual connection from the previous block such that output of each individual block is $\it{LayerNorm}(z + module(z))$. Each Transformer block is computed as follows:

\begin{equation}\begin{aligned}
\mathcal{A}_{M}(\boldsymbol{Q}, \boldsymbol{K}, \boldsymbol{V}) &=\operatorname {concat}(\boldsymbol{h}_{1},\ldots, \text{ $\boldsymbol{h}_{k}$ }) \boldsymbol{W}^{h}, \\
\text { where $\boldsymbol{h}_{i}$ }&= \mathcal{A}\left(\boldsymbol{Q} W_{i}^{Q}, \boldsymbol{K} W_{i}^{K}, \boldsymbol{V} W_{i}^{V}\right) \\
\mathcal{A}(\boldsymbol{Q}, \boldsymbol{K}, \boldsymbol{V})&=\operatorname{softmax}\left(\frac{\boldsymbol{Q} K^{T}}{\sqrt{d_{k}}}\right) \boldsymbol{V}
\label{transformer}
\end{aligned}\end{equation}
The Attention output ($\mathcal{A}$) is a weighted sum of the values ($V$) calculated using a similarity between projected Query ($Q$) and Key ($K$). $\mathcal{A}_{M}$ represents stacked Attention blocks ($\mathcal{A}$), and $W_{i}^{Q}, W_{i}^{K}, W_{i}^{V}$ and $W^{h}$ are parameters to be learnt. 

We utilize Equation \ref{transformer} to perform cross attention between visual spatial representation (RGB $f_{r}$ or Depth $f_{d}$) and language features ($q_{t}$) successively. We do this by utilising the sum of language features and sinusoidal Positional Encoding (PE~\cite{NIPS2017_7181}) as query ($Q = q_{t}$ + PE$(\boldsymbol{q}_{t})$) and visual representation as Key ($K_r = \boldsymbol{f}_{r}$ or $K_d = \boldsymbol{f}_{d}$) as well as Value ($V_r = \boldsymbol{f}_{r}$ or $V_d = \boldsymbol{f}_{d}$).
The final outputs, which we denote as cross-attended context (from RGB or Depth), are computed using $\mathcal{A}_{M}(\boldsymbol{Q}, \boldsymbol{K}, \boldsymbol{V})$, \textit{e.g.,} $\hat{\boldsymbol{I}}^{att}_{q}$ for RGB input and $\hat{\boldsymbol{d}}^{att}_{q}$ for Depth input.

These cross-attended contexts represent the matching between instructions and corresponding visual features at each time step $t$.
Note that the learnable weights in the Transformer are not shared between the two modalities. 

\textbf{Multi-Modal Attention Decoder.}
To decide on which direction to go next and select the most optimal high-level action ($a^{h}_{t}$) high-level policy preserves a temporal memory of the attended visual-linguistic contexts ($\hat{\boldsymbol{I}}^{att}_{q}$, $\hat{\boldsymbol{d}}^{att}_{q}$), mean-pooled visual features ($\hat{\mathrm{v}}_{t}$) and previous actions (${a}^{h}_{t-1}$). We rely on a Recurrent Neural Network to preserve this temporal information across time.
\begin{equation}
\begin{aligned}
\boldsymbol{h}^{h}_{t}&=\operatorname{LSTM}\left(\left[\hat{\boldsymbol{I}}^{att}_{q}, \hat{\boldsymbol{d}}^{att}_{q}, \hat{\boldsymbol{\mathrm{v}}}_{t}, {a}_{t-1}, \boldsymbol{h}^{h}_{t-1}\right]\right) \\
\hat{\boldsymbol{\mathrm{v}}}_{t}&=\boldsymbol{W}_{i}(\mathrm{g}( [\boldsymbol{f}_{r},\boldsymbol{f}_{d}])^\top+\boldsymbol{b}_{i})
\label{recurrence}
\end{aligned}
\end{equation}

where $\mathrm{g(.)}$ is mean adaptive pooling across the spatial dimensions. $W_{i}$ and $b_{i}$ are learned parameters of a fully-connected layer.

The agent computes a probability ($p^{h}_{a}$) of selecting the most optimal action ($a_{t}$) at each time-step by employing a feed-forward network followed by a $softmax$ as follows:
\begin{equation} \begin{aligned}
p^{h}_{a} &= softmax( \boldsymbol{W}_{a}([ \boldsymbol{h}^{h}_{t}] + \boldsymbol{b}_{a}))
\end{aligned}\end{equation}
where $W_{a}$ and $b_{a}$ are parameters to be learnt. High-level action $a_{t}$ comprises of the following navigable directions: \texttt{move forward} (0.25m), \texttt{turn-left} or \texttt{turn-right} (15 degrees) and \texttt{stop}.

\subsection{Low-level Policy}
We employ an imitation policy for the low-level module. 
At each time-step $t$, the low-level policy ($\pi^{l}_{\phi}$) selects a low-level action ($a^{l}_{t})$) given the sub-goal ($a^{h}_{t}$), generated by the high-level policy and observed visual states ($r_{t}, d_{t}$) from the environment. Low-level actions are comprised of agent's linear and angular velocity ($v_{t}, \omega_{t}$). Similar to the high-level module, we use mean pooled visual features ($\hat{\mathrm{v}}_{t}$) for the low-level policy and additionally condition the policy on the high-level sub-goal ($a^{h}_{t}$). Furthermore, we utilize stacked LSTM layers with respective fully-connected layers to generate both low-level action and stop probabilities ($p^{l}_{a},p^{s}_{a}$):
\begin{equation}
\boldsymbol{h}^{l}_{t}=\operatorname{LSTM}\left(\left[\hat{\boldsymbol{v}}_{t}, \boldsymbol{a}^{h}_{t}, \boldsymbol{h}^{l}_{t-1}\right]\right)
\end{equation}
\begin{equation}
p^{h}_{a} = tanh(\mathrm{g_{a}} ([ \boldsymbol{h}^{l}_{t}, \boldsymbol{a}^{l}_{t-1} ])), \quad
p^{s}_{a} = \sigma(\mathrm{g_{s}} ([ \boldsymbol{h}^{l}_{t}, \boldsymbol{a}^{l}_{t-1}]))
\end{equation} 
where $\mathrm{g_{a}(.)}$ and $\mathrm{g_{s}(.)}$ are one-layer Multi-Layer Perceptrons (MLP). $\sigma$ and $tanh$ are sigmoid and tanh activation functions respectively.

\input{tables/comparison}

\input{tables/ablation}

\subsection{Training Details}
We train both high- and low-level policies jointly with three different losses. We employ a multi-class cross-entropy loss computed between ground-truth high-level navigable action ($y^{a}_{t}$) and the predicted action probability ($p^{h}_{a}$) for the high-level policy. We employ a mean squared error loss between ground-truth velocity commands ($y^{v,\omega}_{t}$) and predicted low-level action probabilities ($p^{l}_{a}$). Lastly, we use a binary cross-entropy loss between ground-truth stopping actions ($y^{s}_{t}$) and predicted stop probabilities ($p^{s}_{a}$) as follows:

\begin{multline*}
\begin{gathered}
\mathcal{L}_{\text {loss}}=\lambda \overbrace{\sum_{t=1}^{T} y_{t}^{a} \log \left(p^{a}_{h}\right)}^{\text {High-Level Action Loss}} + (1-\lambda) (\overbrace{\sum_{t=1}^{T}\left(y_{t}^{v,\omega}-p^{l}_{a}\right)^{2}}^{\text{Low-Level Action Loss }} \\ + \overbrace{\sum_{t=1}^{T} y^{s}_{t} \log \left(p^{s}_{a}\right)}^{\text{Low-Level Stop Loss }})
\end{gathered}
\end{multline*}

\section{Dataset and Implementations}
\textbf{Simulation and Dataset.}
We use Habitat simulator~\cite{habitat19iccv} to perform our experiments. Our dataset, Robo-VLN, is built upon Matterport3D dataset~\cite{Matterport3D}, which is a collection of 90 environments captured through around 10k high-definition RGB-D panoramas.
Robo-VLN provides 3177 trajectories, and each trajectory is associated with 3 instructions annotated by humans ported over from the R2R Dataset~\cite{mattersim}. Overall, the dataset comprises 9533 expert instruction-trajectory pairs with an average trajectory length of 326 steps compared to 55.8 in VLN-CE~\cite{krantz2020navgraph} and 5 in R2R~\cite{mattersim}. The corresponding dataset is divided into train, validation seen and validation unseen splits.

\textbf{Evaluation Metrics.} We evaluate our experiments on the following key standard metrics described by Anderson et al.~\cite{DBLP:journals/corr/abs-1807-06757} and Gabriel et al.~\cite{49206}: Success rate ($\textbf{SR}$), Success weighted by path length ($\textbf{SPL}$), Normalized Dynamic Time Warping ($\textbf{NDTW}$), Trajectory Length ($\textbf{TL}$) and Navigation Error ($\textbf{NE}$). We use SPL and NDTW as the primary metrics for comparison. Both of these metrics measure the deviation from ground-truth trajectories; SPL places more emphasis on reaching the goal location, whereas NDTW emphasises on following the complete ground-truth path. 

\textbf{Implementation Details.}
We use pre-trained ResNet-50 on ImageNet~\cite{7780459} and pre-trained ConvNet on a large scale point-goal navigation task, DDPPO~\cite{wijmans2020ddppo} to extract spatial features for images and depth modalities successively.
For transformer module, we use a hidden size ($H = 256$), number of Transformer heads ($n_{h}=4$), and the size of feed-forward layer $(FF=1024)$. We found that truncated backpropagation through time~\cite{sutskever2013training} was invaluable to train longer sequence recurrent networks in our case. We used a truncation length of 100 to train attention decoders in both policies. We trained the network for 20 epochs and performed early stopping based on the performance of the model on validation seen dataset.

\begin{figure}[!b]
\centering
\includegraphics[width=\columnwidth]{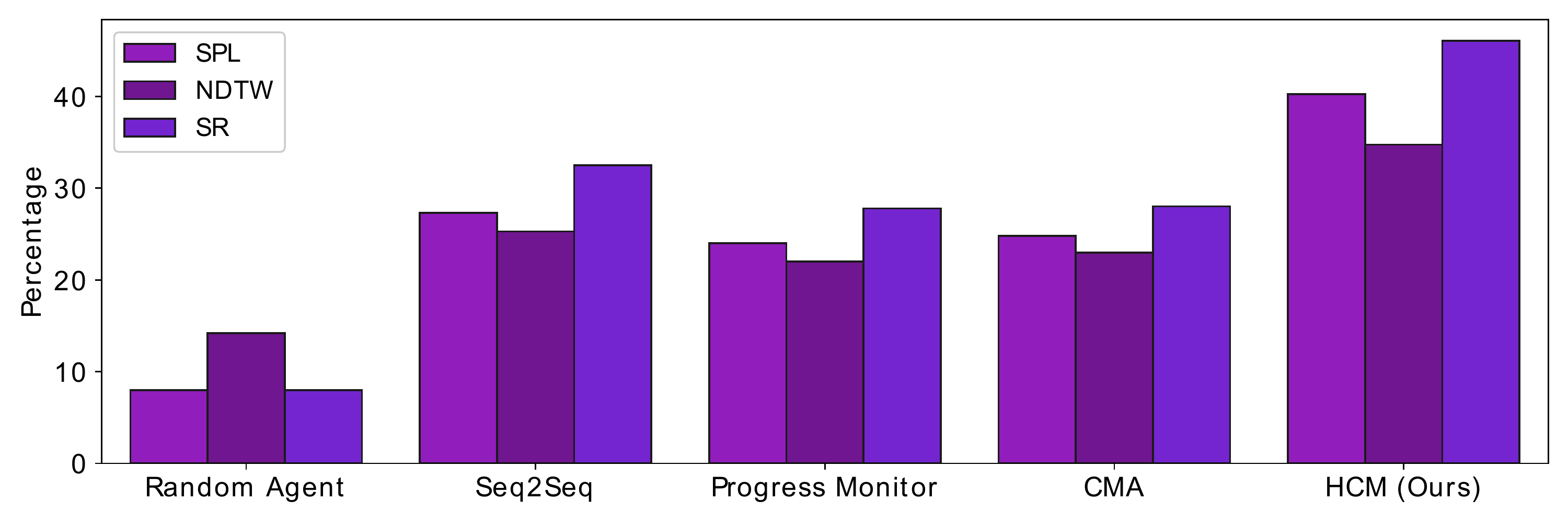}
\centering
  \caption{\textbf{Comparison with strong flat baselines}: Our proposed hierarchical method in comparison with strong flat baselines evaluated on the validation unseen dataset. Our approach shows superior performance and better generalization in unseen settings.
  }
  \label{comparison}
\end{figure}

\begin{figure*}[htp]
\centering
\includegraphics[width=0.9\textwidth]{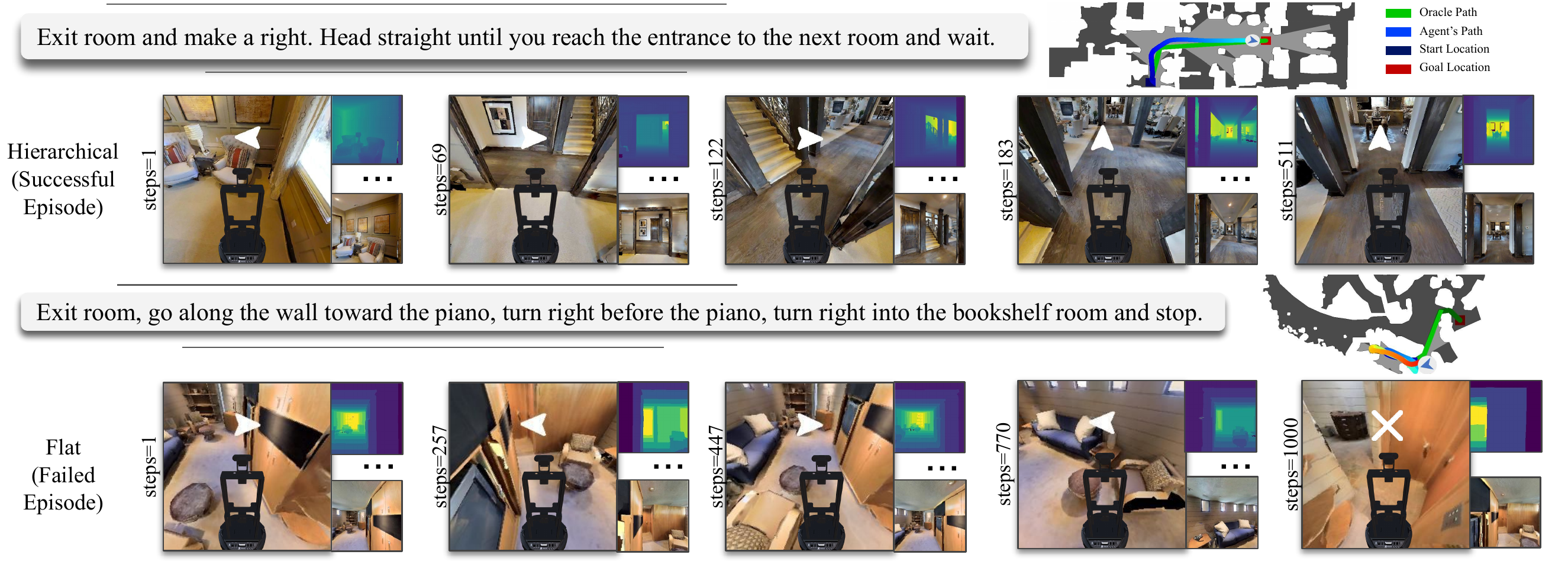}
\captionof{figure}{
\textbf{Qualitative Comparison:} Inference performance of hierarchical and flat model in unseen environments within Robo-VLN. The hierarchical model successfully predicts low-level velocity commands to reach a goal location whereas flat model bumps into obstacles.
}
\label{qualitative}
\end{figure*}

\section{Experiments \& Results} \label{experiments}
\textbf{Flat Baselines.}\label{baselines}
We introduce a suite of flat\footnote{Flat as in there is no explicit hierarchical design for agent's decision making of high- or low-level actions.} baselines that are similar to the ones used in VLN-CE~\cite{krantz2020navgraph}:
(1) \textbf{Sequence-to-Sequence (Seq2Seq)}: an encoder-decoder architecture trained using teacher-forcing~\cite{mattersim},
(2) \textbf{Progress Monitor (PM)}: an agent based on the Seq2Seq model but with an auxiliary loss for progress monitoring, conceptually similar to \cite{ma2019selfmonitoring}, and 
(3) \textbf{Cross-Modal Attention (CMA)}: an cross-modality attention based agent that is conceptually similar to RCM~\cite{wang2019reinforced}.
We adapt these baselines into our Robo-VLN task but with a single change: the output layers now predicts linear and angular velocities as well as the stop action, as opposed to the four actions (forward, turn-left, turn-right, and stop) used in VLN-CE.
Note that baselines are without DAgger~\cite{ross2011reduction} and data augmentation from~\cite{tan2019learning}.

\textbf{Comparison with Flat Baselines.}
The results of our proposed HCM against baselines are summarized in Table~\ref{comparison_table}. As shown in Table~\ref{comparison_table} and Figure~\ref{comparison}, 
our proposed approach, which uses a hierarchical structure to tackle the long-horizon Robo-VLN problem, consistently outperforms the strong baseline models.  
Specifically, our HCM agent shows superior validation unseen performance by achieving a 40\% SPL and 46\% SR; hence demonstrating an absolute 13\% improvement in SR and 10\% improvement in SPL over the best performing baseline on the validation unseen environments.  

\textbf{Ablation Study.}
In our ablation experiments, we empirically validate the significance of different design choices and modules in our proposed HCM agent. Our results are summarized in~Table~\ref{ablation}. First, we ablate \textit{vision} (RGB and Depth) in our model. Our results show that an agent without vision performs as good as a random agent (\textit{i.e.,} 0.07 SPL, 0.07SR). It shows the effectiveness of vision for end-to-end trainable agents in photo-realistic simulations. Second, we consider an architecture with early RGB and Depth fusion before cross attention with language. Our results show that separately aligning RGB and Depth with instructions performs much better than attending to the instructions corresponding to a fused RGB-D representation. We further ablate \textit{hierarchy} to show the importance of hierarchy in our architecture. Our results are summarized as follows.

\textbf{Is the source of improvement from \textit{hierarchy}?}
Our method relies on decomposing the complex task into layered decision making; the top level predicts a sub-goal whereas the bottom level predicts low-level velocity commands.
To confirm that hierarchy is indeed the source of improvement, we devise an experiment, in which we \textit{flattened} the hierarchical model and provide auxiliary sub-goal supervision to the flattened model in addition to the low-level supervisions.
This model effectively reduced to Seq2Seq baseline model but with high-level action supervision. 
The results are reported in Table~\ref{ablation} (\#2 vs \#4). 
We show that, despite using same levels of supervisions, the flattened hierarchical model under-performs the hierarchical approach, \textit{e.g.,} 40\% vs 46\% in SR and 34\% vs 40\% in SPL. 
This comparison demonstrates that decoupling reasoning and imitation indeed plays a pivotal role in learning effective individual policies.

\textbf{Qualitative Comparison.}
We qualitatively analyze the performance of hierarchical and flat agents in Robo-VLN. As shown in Figure \ref{qualitative}, the hierarchical agent (top example) successfully predicts low-level velocity commands while reaching a desired goal location described by the instruction. The agent takes significantly more steps than discrete VLN settings (511 steps) to reach the goal location; hence showing the effectiveness of hierarchical agents to solve long horizon cross-modal trajectory following problem. The flat agent (bottom figure) fails to follow the trajectory and drives into obstacles multiple times. The episode ends after the agent is unsuccessful in reaching the goal at 1000 steps.

\section{conclusion} \label{sec:5}
Despite the recent progress, existing VLN environments impose certain unrealistic assumptions such as perfect localization, known topology and deterministic navigation in the absence of any obstacles. 
In this work, we first propose the Robo-VLN setting that lifts off the unrealistic assumption of navigation graph and discrete action space and provides a suite of strong baselines inspired by the recent works in discrete VLN setting. We then take the next step to propose a Hierarchical Cross-Modal (HCM) agent that tackles the challenging long-horizon issue in Robo-VLN via a hierarchical model design. Our proposed HCM agent, with trained high- and low-level policies, achieves significant performance improvement against the strong baselines. We believe that our new Robo-VLN setting and strong benchmarks would help build a stronger suite of autonomous agents.
\bibliography{bibliography.bib}
\end{document}

%% file: tables/comparison.tex



\begin{table*}[t]
    \scriptsize
    \centering
    \renewcommand{\arraystretch}{1.3}
    \vspace{+0.15cm}
    \caption{
    \textbf{Quantitative comparison}: Comparison with strong baselines.
    Note that these baselines are reimplementations from VLN-CE~\cite{krantz2020navgraph} with small changes (see Section~\ref{baselines} for further details). 
    }
    \label{comparison_table}
    \resizebox{0.9\textwidth}{!}{
    \begin{tabular}{cccccccccccc}
        \toprule
        & & \multicolumn{5}{c}{\textbf{Validation Seen}} & \multicolumn{5}{c}{\textbf{Validation Unseen}} \\ \cline{3-12}
        & {Method} & \textbf{SR}~$\uparrow$ & \textbf{SPL}~$\uparrow$ & \textbf{NDTW}~$\uparrow$ & \textbf{TL}~$\uparrow$ & \textbf{NE}~$\downarrow$ & \textbf{SR}~$\uparrow$ & \textbf{SPL}~$\uparrow$ & \textbf{NDTW}~$\uparrow$ & \textbf{TL}~$\uparrow$ & \textbf{NE}~$\downarrow$ \\
        \midrule
        1 & {Random Agent} &  0.07& 0.07 & 0.14  & 5.26 & 10.25 & 0.08 & 0.08& 0.14  & 5.40 & 9.81 \\
        2 & {Seq2Seq~\cite{Matterport3D}} &0.36	&0.34&	0.32&	11.84&	8.63& 0.33	&	0.30	&0.28&	11.92&	8.97 \\
        3 & {PM~\cite{ma2019selfmonitoring}} & 0.32 & 0.27 & 0.23 & 14.12 & 9.33 & 0.28 & 0.24 & 0.22 &  13.85 & 9.82\\
        4 & {CMA~\cite{wang2019reinforced}} & 0.28&	0.25&	0.22&	11.52&	9.95&	0.28&	0.25&	0.23&	11.57&	9.63 \\
        \midrule
         & {\textbf{HCM (Ours)}} & \textbf{0.49}&	\textbf{0.43}&	\textbf{0.35}	& \textbf{13.53} &	\textbf{7.48} & \textbf{0.46}&	\textbf{0.40} &	\textbf{0.35} & \textbf{14.06} &	\textbf{7.94} \\
        \bottomrule
    \end{tabular}
    }
\end{table*}

%% file: tables/ablation.tex


\begin{table*}[t]
    \centering
    \renewcommand{\arraystretch}{1.3}
    \caption{
    \textbf{Ablation Study}: Impact of different modules and design choices in our proposed Hierarchical agent.
    }
    \label{ablation}
    \resizebox{0.9\textwidth}{!}
{
    \begin{tabular}{cccccccccccccccc}
        \toprule
        & & \multicolumn{3}{c}{\textbf{Module}}& \multicolumn{5}{c}{\textbf{Validation Seen}} &\multicolumn{5}{c}{\textbf{Validation Unseen}} \\ \cline{3-15}  
& \# & {Vision} &{Hierarchy} & \begin{tabular}[c]{@{}c@{}}RGB-D\\ Early fusion\end{tabular} & \textbf{SR}~$\uparrow$ & \textbf{SPL}~$\uparrow$ & \textbf{NDTW}~$\uparrow$ & \textbf{TL}~$\uparrow$ & \textbf{NE}~$\downarrow\uparrow$ & \textbf{SR}~$\uparrow$ & \textbf{SPL} & \textbf{NDTW}~$\uparrow$ & \textbf{TL}~$\uparrow$ & \textbf{NE}~$\downarrow$ \\
\midrule
\multirow{4}{*}{\shortstack{Hierarchical\\Agent}} & 1 & &\checkmark &  & 0.07 & 0.07& 0.14& 4.82 & 10.34 & 0.07& 0.07 & 0.14 & 10.2 & 4.81  \\
& 2 &\checkmark& & & 0.44& 0.37 & 0.31& 14.87 & 8.21 & 0.40 & 0.34 & 0.28 &  15.32 & 8.64  \\
& 3 &\checkmark& \checkmark& \checkmark & 0.39 &  0.35 & 0.29 & 13.87 & 9.13 & 0.34 & 0.31 & 0.28 & 12.85 & 8.78  \\
& 4 &\checkmark&\checkmark & & 0.49&	0.43&	0.35	&13.53&	7.48&	0.46&	0.40&	0.35&	14.06&	7.94  \\
\bottomrule
    \end{tabular}
}
\end{table*}